# ZERO-SHOT AUDIO CLASSIFICATION BASED ON CLASS LABEL EMBEDDINGS

*Huang Xie*\*

Tampere University
Tampere, 33720, Finland
huang.xie@tuni.fi

*Tuomas Virtanen*\*

Tampere University
Tampere, 33720, Finland
tuomas.virtanen@tuni.fi

## ABSTRACT

This paper proposes a zero-shot learning approach for audio classification based on the textual information about class labels without any audio samples from target classes. We propose an audio classification system built on the bilinear model, which takes audio feature embeddings and semantic class label embeddings as input, and measures the compatibility between an audio feature embedding and a class label embedding. We use VGGish to extract audio feature embeddings from audio recordings. We treat textual labels as semantic side information of audio classes, and use Word2Vec to generate class label embeddings. Results on the ESC-50 dataset show that the proposed system can perform zero-shot audio classification with small training dataset. It can achieve accuracy (26 % on average) better than random guess (10 %) on each audio category. Particularly, it reaches up to 39.7 % for the category of natural audio classes.

## 1. INTRODUCTION

Supervised audio classification requires large amounts of annotated data of target classes to train models to achieve relatively good performance. However, existing audio datasets are limited to certain predefined sound classes, such as AudioSet [1] (roughly 2 million human-labeled audio samples of 527 classes), Freesound [2] (roughly 300,000 audio samples of over 600 classes), ESC-50 [3] (2000 audio samples of 50 classes), etc. On the other hand, addressing new audio recognition tasks would require collecting new dataset with annotated target classes, which is time consuming and costly.

In recent years, zero-shot learning has received increasing attention. The term, zero-shot learning, is first coined in [4]. For zero-shot learning problems, there is annotated data from some training classes, but from the target classes only semantic side information is available. Based on existing audio datasets [1, 2, 3], there is potential in using zero-shot learning to obtain classification models for new classes.

Recently, zero-shot learning has been investigated in the context of computer vision [5]. Most of these approaches try to overcome zero-shot learning problems by means of leveraging intermediate-level semantic representations for both samples and their classes [6, 7, 8]. It is generally assumed that these representations are shared between the source domain (training data) and target domain (testing data). Then, a model is established to learn the relationship between samples and classes in the intermediate-level representation spaces. The commonly used semantic representations of classes include attribute embeddings [7, 8] and word vectors [9]. However, most of these existing approaches are mainly used for image recognition. In contrast, there is no research work has been conducted for zero-shot learning on audio classification task.

In this paper, we propose a zero-shot learning approach for audio classification. First, we use VGGish [10] to extract audio feature embeddings from audio recordings, and generate semantic class label embeddings from the textual labels of audio classes with Word2Vec [9]. Then, we employ a bilinear model [7] to learn the relationship between audio feature embeddings and class label embeddings. Experimental results on the ESC-50 dataset show that the proposed method can be used to perform zero-shot audio classification with a small dataset. In comparison to random guess (10 %), the average classification accuracy reaches around 26 % on all the testing audio categories.

The remainder of this paper is organized as follows: first, we give a detail description of our approach to conduct zero-shot learning for audio classification in Sec. 2. Then, we introduce the settings of our experiments and report the results in Sec. 3. Finally, we conclude this paper in Sec. 4.

## 2. PROPOSED METHOD

In this section, we introduce our zero-shot learning approach for audio classification. The block diagram of the overall approach is illustrated in Figure 1. First, we use VGGish [10] to extract audio feature embeddings from audio recordings and generate semantic class label embeddings from the textual labels of audio classes with Word2Vec [9]. We do audio classification based on the bilinear model [7]. It takes audio feature embeddings and semantic class label embeddings as input, and is trained to measure the compatibility between an audio feature embedding and a class label embedding. The classification output is the class that has the maximum compatibility with the given audio feature embedding as output.

In what follows, we first describe how to extract audio feature embeddings with VGGish [10], and we also discuss

---

\* The research leading to these results has received funding from the European Research Council under the European Unions H2020 Framework Programme through ERC Grant Agreement 637422 EVERYSOUND.

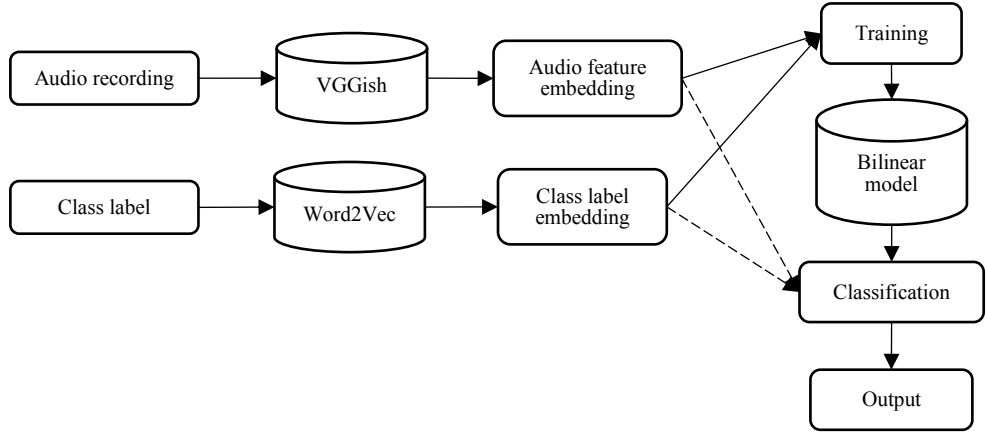

Figure 1. Proposed framework for zero-shot audio classification

Word2Vec [9] for generating semantic class label embeddings from the textual class labels. Then, we explain how to incorporate these embeddings into the bilinear model [7].

### 2.1. Audio Feature Embedding

VGGish is a variant of the VGG model [10], and it can be used to generate high abstraction level feature embeddings from audio recordings. In previous work [10], it is shown that a classification model can achieve better performance with VGGish generated embeddings than raw audio features on an audio classification task.

In our approach, we follow [10] and use VGGish to generate audio feature embeddings from audio recordings. An input audio recording is first resampled to 16 kHz mono. Then, a log mel spectrogram tensor (96×64) is extracted from every one-second segment. VGGish takes the log mel spectrogram tensors as input. For one second of audio, it produces one embedding vector. To present the whole audio signal, we take the average of its one-second embedding vectors. We use the VGGish model that is pre-trained with YouTube-8M and published by Google [10].

### 2.2. Class Label Embedding

It is generally assumed that class labels convey semantic side information about classes. To take into account this side information in a classification task, one of the commonly used approaches is to transform them into high-dimensional vector space and obtain vector representations for the corresponding classes.

Word2Vec is an efficient model for learning high-level vector representations of words and phrases [9]. In our approach, we use it to create numerical representations for class labels.

For the sake of simplicity, we use the pre-trained Word2Vec model published by Google [9]. It contains 300-dimensional vector representations for about 3 million words and phrases. As this pre-trained model is case-sensitive, we cast the textual label into lowercase. For the single-word labels, we use the semantic word vector as the class label embedding for a class. For multi-word labels, a class label embedding is obtained by calculating the average of the individual semantic word vectors.

**Training algorithm 1**:
1: Initialize $\mathcal{W}$ with zeros.
2: For epoch = 1 to $m$ do
3:   For each $(x_n, y_n) \in S$ do
4:     Calculate $\mathcal{L} = \{\ell(x_n, y_n, y; \mathcal{W}) \mid y \in \mathcal{Y}\}$
5:     Let $\mathcal{L}' = \langle \ell_r \rangle_{r=1}^C$ be the sequence of sorted elements in $\mathcal{L}$
6:     For each $\ell_r(x_n, y_n, y; \mathcal{W})$ in $\mathcal{L}'$ do
7:       If $\ell_r(x_n, y_n, y; \mathcal{W}) > 0$ then
8:         $\mathcal{W} = \mathcal{W} - \eta \cdot \beta_{\lfloor (C-1)/r \rfloor} \cdot \theta(x_n) \cdot [\varphi(y) - \varphi(y_n)]^T$
9:       End if
10:    End for
11:  End for
12: End for

### 2.3. Bilinear Model

The bilinear model, proposed in [7], provides a joint embedding framework, which allows to conduct zero-shot learning with feature embeddings of sample $x \in \mathcal{X}$ and class $y \in \mathcal{Y}$.

Given the parameter matrix $\mathcal{W}$, it defines a compatibility function $\mathcal{F} : \mathcal{X} \times \mathcal{Y} \to \mathbb{R}$ to measure the compatibility between $x$ and $y$:

$$\mathcal{F}(x, y; \mathcal{W}) = \theta(x)^T \cdot \mathcal{W} \cdot \varphi(y) \quad (1)$$

where $\theta(x)$ and $\varphi(y)$ denote the feature embeddings of sample $x$ and class $y$, respectively. For our audio classification task, we define $\theta(x)$ as audio feature embedding, and $\varphi(y)$ as class label embedding.

For prediction, we use function $f(x; \mathcal{W})$ to produce the target class:

$$f(x; \mathcal{W}) = \underset{y \in \mathcal{Y}}{\operatorname{argmax}} \mathcal{F}(x, y; \mathcal{W}) \quad (2)$$

**Training**. Given a training set $S = \{(x_n, y_n) \mid x_n \in \mathcal{X}, y_n \in \mathcal{Y}, n = 1, \dots, N\}$, the bilinear model can be trained through stochastic gradient descent (SGD). Here, we use the similar learning algorithm as proposed in [7, 8] to train and learn the parameter matrix $\mathcal{W}$. In the **training algorithm 1**, $\eta$ is the learning rate, and $C$ is the number of classes. Let $r$ be the loss rank of a class $y$ for a given training sample $(x_n, y_n)$. With the ranking penalty $\beta_r$, we consider all loss values in a weighted manner. $\beta_r$ is defined as:

$$\beta_r = \sum_{j=1}^{r} \frac{1}{j} \quad (3)$$

Our main goal is to learn a parameter matrix $\mathcal{W}$ by minimizing the following empirical risk for $S$:

$$\frac{1}{N} \sum_{n=1}^{N} \frac{\beta_{r(x_n,y_n;\mathcal{W})}}{r(x_n,y_n;\mathcal{W})} \sum_{y \in \mathcal{Y}} max\{0, \ell(x_n, y_n, y; \mathcal{W})\} \quad (4)$$

where $r(x_n, y_n; \mathcal{W})$ is the rank of class $y_n$ for sample $x_n$. It defines a weighted ranking loss with the convention $0/0 = 0$ when the correct class is top-ranked [11].

The loss function $\ell(x_n, y_n, y; \mathcal{W})$ is defined as follows:

$$\ell(x_n, y_n, y; \mathcal{W}) = \Delta(y_n, y) + \mathcal{F}(x_n, y; \mathcal{W}) - \mathcal{F}(x_n, y_n; \mathcal{W}) \quad (5)$$

where $\Delta(y_n, y) = 0$ if $y_n = y$ and 1 otherwise. It consists of two components. The $\Delta(y_n, y)$ is the target loss to measure classification error, and the rest part acts as the surrogate loss at learning time. By minimizing (4) and (5), the correct class of a given sample would be top-ranked.

## 3. EXPERIMENTS

In this section, we describe settings and results of our experiments.

### 3.1. Dataset

We conduct our experiments based on the public audio dataset ESC-50 [4]. It contains 2000 labeled 5-second-long audio recordings of 50 classes (with 40 samples per class), and these 50 classes are arranged into 5 major audio categories (with 10 classes per category): animals, natural soundscapes & water sounds, human & non-speech sounds, interior sounds, exterior noises. The class labels in each audio category are shown in Table 1.

### 3.2. Set-up

In our experiments, we evaluate our audio classification system with zero-shot learning settings and measure the classification accuracy on the ESC-50 classes. In each setting, the dataset is split into training and testing class wise, and the test classes are not used for training. In addition to zero-shot learning, we also conduct a setting for few-shot learning where a small number of samples from the target class are used in training. We partition the ESC-50 dataset with two strategies: category-based and class-based.

#### 3.2.1. Category-based strategy

In the category-base strategy, we use classes of one category to learn the parameter matrix $\mathcal{W}$, and evaluate the system with classes of one other category.

**Setting 1**. For each category, we apply 5-fold cross-validation. The 10 classes are grouped into 5 folds (with 2 classes per fold) at random. Then, 8 classes (320 samples) are used for training the parameter matrix $\mathcal{W}$, and the remaining 2 classes (80 samples) for evaluation. We compare the classification accuracy on different categories in Table 2. Our results show our audio classification system is capable of performing zero-shot learning with small dataset. On average, it achieves better classification accuracy than random guess (50 %) on each category.

| Category | Class labels |
|---|---|
| Animals | dog, rooster, pig, cow, frog, cat, hen, insects, sheep, crow |
| Natural | rain, sea waves, crackling fire, crickets, chirping birds, water drops, wind, pouring water, toilet flush, thunderstorm |
| Human | crying baby, sneezing, clapping, breathing, coughing, footsteps, laughing, brushing teeth, snoring, drinking sipping |
| Interior | door wood knock, mouse click, keyboard typing, door wood creaks, can opening, washing machine, vacuum cleaner, clock alarm, clock tick, glass breaking |
| Exterior | helicopter, chainsaw, siren, car horn, engine, train, church bells, airplane, fireworks, hand saw |

Table 1: Class labels of different audio categories

**Setting 2**. We use 10 classes (400 samples) of one category to train the classification system, and 10 classes (400 samples) of another category for evaluation. In total, there are 20 training-test category pairs. As shown in Table 2, for all these categories, the system performs a little better than random guess (10 %). We also notice that our system is not robust to different training categories. The variation of accuracy can reach up to 13.5% on the same evaluation category. We can conclude that the bilinear model used to build the classification system seems to be sensitive to the training settings. With different training classes, its performance can vary dramatically.

#### 3.2.2. Class-based strategy

Based on the experiment results from category-based strategy, we combine classes from different categories to form a training dataset and evaluate the model with the remaining classes.

**Setting 3**. First, we combine four categories and use their 40 classes (1600 samples) for training. The 10 classes (400 samples) of the remaining category are used for evaluation. Results are shown in Table 4. Compared with Setting 2, the classification system achieves better accuracy on each category with more training classes and samples.

| Category | Top-1 Accuracy (%) |
|---|---|
| Animals | 60.2 |
| Natural | 83.0 |
| Human | 75.5 |
| Interior | **88.5** |
| Exterior | 74.7 |

Table 2: Classification accuracy on different categories

| Training Category | Top-1 Accuracy (%) | | | | |
|---|---|---|---|---|---|
| | Animal | Natural | Human | Interior | Exterior |
| Animal | - | 21.0 | 21.5 | 16.7 | 15.5 |
| Natural | 14.2 | - | 15.7 | 14.2 | 26.2 |
| Human | 16.2 | 14.5 | - | 17.7 | 17.5 |
| Interior | 15.0 | 9.7 | 11.2 | - | 24.7 |
| Exterior | 14.0 | 23.5 | 14.7 | 13.0 | - |

Table 3: Classification accuracy with different training-test category pairs

| Evaluation Category | Top-1 Accuracy (%) | | |
|---|---|---|---|
| | Setting 2 (averaged) | Setting 3 | Setting 4 |
| Animal | 14.9 | 22.2 | 51.2 |
| Natural | 17.2 | **39.7** | 68.4 |
| Human | 15.8 | 22.0 | 56.3 |
| Interior | 15.4 | 17.0 | 58.7 |
| Exterior | 21.0 | 29.2 | 73.8 |

Table 4: Classification accuracy with different training setting

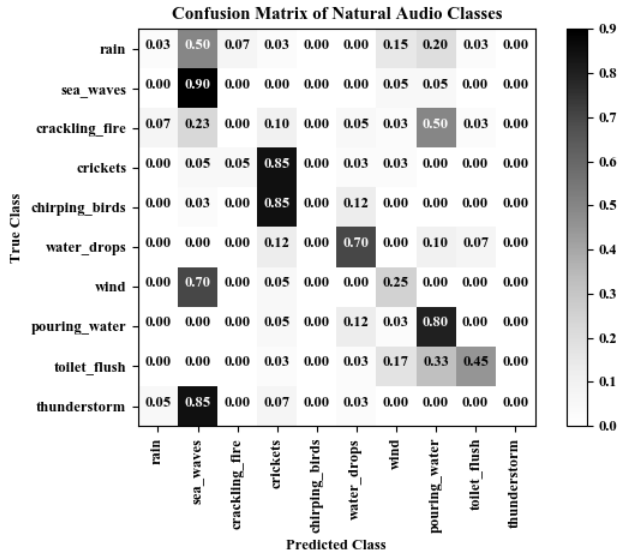

Figure 2. Confusion matrix for natural audio classes

On average, it achieves accuracy around 26 % on all these audio categories. Particularly, the accuracy reaches up to 39.7 % for the category of natural audio classes, and the confusion matrix is shown in Figure 2. We notice that misclassification occurs mainly among classes that are semantically similar. For instance, the class "sea waves" dominates the classification of classes "rain", "sea waves", "wind", "thunderstorm", while "crickets" dominate the classification of "crickets" and "chirping birds" since these classes are semantically more similar.

**Setting 4**. In addition to the above zero-shot learning settings, we conduct Setting 4 for few-shot learning. In these experiments, we have a few training samples from the 10 evaluation classes in addition to the 40 training classes. For each evaluation class, we apply 8-fold cross-validation: the 40 samples are grouped into 8 folds (with 5 samples per fold) at random. Then, 5 samples are added into the training dataset, and the remaining 35 samples as the testing dataset. The experimental results are shown in Table 4 (Setting 4). Compared with Setting 2 and 3, the performance of our classification system is improved significantly by adding a few samples from the evaluation classes, and achieves higher accuracy on each evaluation category.

## 4. CONCLUSIONS

In this paper, we propose a zero-shot learning approach that is able to do classification of audio samples from classes, which have no available samples for training the system but only textual information about the class labels. We use VGGish to extract audio feature embeddings from audio recordings, and generate semantic class label embeddings from the textual labels of audio classes with Word2Vec. We build our audio classification system based on the bilinear model. It takes audio feature embeddings and semantic class label embeddings as input, and is trained to measure the compatibility between an audio feature embedding and a class label embedding. To do classification, it produces the class that has the maximum compatibility with the given audio feature embedding as output.

In our experiments, we use ESC-50 dataset and evaluate our system with zero-shot learning settings for audio classification. The experimental results show that our classification system is capable of performing zero-shot learning with small dataset, and it can achieve accuracy (26 % on average) better than random guess (10 %). Particularly, the accuracy reaches up to 39.7 % for the category of natural audio classes.

As it is sensitive to the training settings, the performance of classification can vary dramatically. By increasing the volume of training data, it can achieve higher classification accuracy. Further, we feel it is necessary to explore its performance on larger audio datasets, such as AudioSet.

## 5. REFERENCES

[1] J. Gemmeke, D. Ellis, D. Freedman, et al. Audio Set: An ontology and human-labeled dataset for audio events. In ICASSP, 2017.

[2] E. Fonseca, J. Pons, X. Favory, F. Font, et al. Freesound datasets: a platform for the creation of open audio datasets. In ISMIR, 2017.

[3] K. J. Piczak. ESC: Dataset for environmental sound classification. In Proceedings of the 23rd ACM international conference on Multimedia. ACM, 2015, pp. 1015-1018.

[4] M. Palatucci, D. Pomerleau, G. Hinton, and T. Mitchell. Zero-shot learning with semantic output codes. In NIPS, 2009.

[5] Y. Fu, T. Xiang, Y. Jiang, X. Xue, L. Sigal, S. Gong. Recent advances in zero-shot recognition. In CVPR, 2017.

[6] B. Romera-Paredes, P. Torr. An embarrassingly simple approach to zero-shot learning. In ICML, 2015.

[7] Z. Akata, F, Perronnin, Z. Harchaoui, and C. Schmid. Label-embedding for image classification. In CVPR, 2015.

[8] Y. Xian, Z. Akata, G. Sharma, Q. Nguyen, M. Hein, and B. Schiele. Latent embeddings for zero-shot classification. In CVPR, 2016.

[9] T. Mikolov, I. Sutskever, K. Chen, G. Corrado, and J. Dean. Distributed representations of words and phrases and their compositionality. In NIPS, 2013.

[10] S. Hershey, S. Chaudhuri, D. Ellis, J. Gemmeke, et al. CNN architectures for large-scale audio classification. In ICASSP, 2017.

[11] J. Weston, S. Bengio, and N. Usunier. Large scale image annotation: Learning to rank with joint word-image embeddings. In ECML, 2010.